\documentclass{svproc}
\usepackage{mathptmx}
\usepackage{helvet}
\usepackage{courier}
\usepackage{makeidx}
\usepackage{amsmath,amsfonts}
\usepackage{graphicx}
\usepackage{cite}
\usepackage[bottom]{footmisc}
\usepackage{bm}

\begin{document}

\mainmatter

\title{Geometric Interpolation of Rigid Body Motions}

\titlerunning{Predicting Rigid Body Motions}

\author{Andreas M\"uller\inst{1}}

\authorrunning{Andreas M\"uller}

\tocauthor{Andreas M\"uller}

\institute{Johannes Kepler University Linz, Institute of Robotics\\
Altenberger Str. 69, 4040 Linz, Austria\\
\email{a.mueller@jku.at}}

\maketitle

\abstract{
The problem of interpolating a rigid body motion is to find a spatial
trajectory between a prescribed initial and terminal pose. Two variants of
this interpolation problem are addressed. The first is to find a solution
that satisfies initial conditions on the $k-1$ derivatives of the rigid body
twist. This is called the $k$\emph{th-order initial value trajectory
interpolation problem (}$k$-\emph{IV-TIP)}. The second is to find a solution
that satisfies conditions on the rigid body twist and its $k-1$ derivatives
at the initial and terminal pose. This is called the $k$\emph{th-order
boundary value trajectory interpolation problem (}$k$-\emph{BV-TIP)}.
Solutions to the $k$-IV-TIP for $k=1,\ldots ,4$, i.e. the initial twist and
up to the 4th time derivative are prescribed.  Further, a solution to the
1-IV-TBP is presented, i.e. the initial and terminal twist are prescribed.
The latter is a novel cubic interpolation between two spatial configurations
with given initial and terminal twist. This interpolation is automatically
identical to the minimum acceleration curve when the twists are set to zero.
The general approach to derive higher-order solutions is presented.
Numerical results are shown for two examples.
}

\vspace{-3ex}

\section{Introduction}

Smooth interpolation of spatial rigid body motions through given poses is an
important task in robot motion planning. This gained renewed relevance with
the advent of aerial robots, and UAV in particular. Early approaches to
motion interpolation treated translations and rotations separately. Spatial
rotation interpolation has been the addressed in a series of publications,
for whish the spherical linear interpolation (SLERP) \cite{Shoemake1985} is
now an established method. This method uses a linear interpolation of the
scaled rotation vector, which serves as canonical coordinates. The classical
SLERP represents rotations as unit quaternions forming the Lie group $%
Sp\left( 1\right) $. These canonical coordinates can also be represented by
orthogonal matrices, which form the Lie group $SO\left( 3\right) $. With
this parameterizations, the cubic interpolation of rotations was addressed
in \cite{ParkRavani-JMD1995,ParkRavani1997,KangPark1998}, which admits
prescribing the initial and terminal angular velocity. Realizing that for
general rigid body motions, rotations and translations are coupled, the
motion interpolation problem was formulated on the Lie group $SE\left(
3\right) $ \cite{ZefranKumar1998,ZefranKumarCroke-TRO1998,Selig-IROS2006}.
While the SLERP interpolation of rotations yields a geodesic w.r.t. the
bi-invariant metric on $SO\left( 3\right) $, adopting SLERP to rigid body
motion interpolation yields geodesic w.r.t. the left-invariant metric on $%
SE\left( 3\right) $. Motion trajectories that minimize acceleration and jerk
were investigated in \cite{Noakes1989,ZefranKumarCroke-TRO1998,Selig-IMA2007}%
. In these publications, the differential equations of a boundary value
problem are presented whose solution are the minimum acceleration
trajectories between two given poses. In \cite%
{ParkRavani-JMD1995,ParkRavani1997,KangPark1998}, a cubic interpolation
scheme is derived assuming that the canonical coordinates have a general
cubic dependency on time.

In this paper a two-point interpolation scheme is derived making use of a
third-order approximation of the Poisson-Darboux equation. {The motion of a
rigid body, represented by a body-fixed frame, relative to a space-fixed
inertial frame is described by $\mathbf{C}\left( t\right) \in SE\left(
3\right) $. }The rigid body twist in spatial representation, defined as $%
\dot{\mathbf{C}}=\hat{\mathbf{V}}\mathbf{C}$, is used.

The \emph{geometric interpolation problem} is to find a curve in $SE\left(
3\right) $ that connects a prescribed initial configuration $\mathbf{C}_{0}$
and terminal configuration $\mathbf{C}_{T}$. This curve is parameterized by
a path parameter $t\in \left[ 0,T\right] $ (e.g. time), and $\mathbf{C}_{0}:=%
\mathbf{C}\left( 0\right) $ and $\mathbf{C}_{T}:=\mathbf{C}\left( T\right) $%
. It will be convenient to use the normalized parameter $\tau :=t/T$, such
that $\tau \in \left[ 0,1\right] $. A solution of the geometric
interpolation problem is given by $\mathbf{C}\left( \tau \right) =\exp
\left( \tau \mathbf{X}_{T}\right) \mathbf{C}_{0}$, where $\mathbf{X}%
_{T}=\log (\mathbf{C}_{0}^{-1}\mathbf{C}_{T})$ is the constant screw
coordinate vector. This solution is a geodesic on $SE\left( 3\right) $, i.e.
the motion with shortest path, w.r.t. a (any) left-invariant metric on $%
SE\left( 3\right) $. Notice that $\mathbf{X}_{T}$ is represented in the
inertial frame. The $k$\emph{th-order initial value trajectory interpolation
problem (}$k$-\emph{IV-TIP)} is to find a solution to the geometric
interpolation problem which additionally satisfies initial conditions on the
rigid body twist and its $k-1$ derivatives. The $k$\emph{th-order boundary
value trajectory interpolation problem (}$k$-\emph{BV-TIP)} is to find a
solution to the geometric interpolation problem which additionally satisfies
conditions on the rigid body twist and its $k-1$ derivatives at the initial
and terminal point. First- and second-order interpolations, for example, are
important for motion planning. Solutions to the trajectory interpolation
problem have the general form {$\mathbf{C}$}$\left( \tau \right) =\exp
\left( \mathbf{X}\left( \tau \right) \right) \mathbf{C}_{0}$, with
non-constant $\mathbf{X}$, and are not geodesics anymore. In this paper, the 
$k$-IV-TIP is solved for $k=1,\ldots ,4$, and the first-order trajectory
interpolation problem 1-IV-BIP is solved. The solution is derived as a
third-order interpolation scheme.

For the subsequent derivations, in addition to the normalized time $\tau $,
the normalized $i$th time derivative of the twist is introduced as $\bar{%
\mathbf{V}}^{\left( i\right) }:=T^{i+1}\mathbf{V}^{\left( i\right) }$, where
the notation $\mathbf{V}^{\left( i\right) }:=\frac{d^{i}}{dt^{i}}\mathbf{V}$
is used.

\section{Rigid Body Kinematics --Poisson-Darboux Equation}

The twist in body-fixed representation is defined by the right
Poisson-Darboux equation on $SE\left( 3\right) $%
\begin{equation}
\dot{\mathbf{C}}=\hat{\mathbf{V}}\mathbf{C}  \label{Poisson}
\end{equation}%
where $\mathbf{V}\in {\mathbb{R}}^{6}$ is the rigid body twist in \emph{%
spatial representation}. For given twist, and initial pose, these equations
can be solved to determine the pose of a body. They are therefore referred
to as the \emph{kinematic reconstruction equations}. The solution is
expressed as%
\begin{equation}
\mathbf{C}\left( t\right) =\exp \hat{\mathbf{X}}\left( t\right) \mathbf{C}%
_{0}  \label{C}
\end{equation}%
with initial pose $\mathbf{C}_{0}\in SE\left( 3\right) $, and with $\hat{%
\mathbf{X}}\left( t\right) \in se\left( 3\right) $. The rigid body motion is
parameterized by the instantaneous screw coordinate vector $\mathbf{X}\in {%
\mathbb{R}}^{6}\cong se\left( 3\right) $, which serve as canonical
coordinates, and the trajectory in the Lie algebra $se\left( 3\right) $ is
lifted to the trajectory in the group by the exp map on $SE\left( 3\right) $%
. Other non-canonical coordinates can be used, such as Cayley-Rodrigues
parameters \cite{Selig-IFToMM2007,RSPA2021}. As shown by Magnus \cite%
{Magnus1954}, if $\mathbf{C}$ in (\ref{C}) satisfies the ODE (\ref{Poisson})
on $SE\left( 3\right) $, then $\hat{\mathbf{X}}$ satisfies the first-order
ODE on $se\left( 3\right) $

\begin{equation}
\mathbf{V}=\mathbf{dexp}_{\hat{\mathbf{X}}}\dot{\mathbf{X}}  \label{RecExp}
\end{equation}%
where $\mathbf{dexp}_{\mathbf{X}}:se\left( 3\right) \rightarrow se\left(
3\right) $ is the right-trivialized differential of the exp map. Given the
instantaneous screw coordinate vector $\mathbf{X}$, its time derivative is
related to the spatial twist $\mathbf{V}$ by the inverse relation 
\begin{equation}
\dot{\mathbf{X}}=\mathbf{dexp}_{\hat{\mathbf{X}}}^{-1}\mathbf{V}.
\label{Xdot}
\end{equation}%
The right-trivialized differential on $SE\left( 3\right) $ and its inverse
admit closed form relations. A summary of different expressions can be found
in \cite{RSPA2021}. It was already shown by Hausdorff \cite[pp. 26 \& 36ff]%
{Hausdorff1906} (see \cite{Iserles1984} for a proof) that the
right-trivialized differential and it inverse admit the series expansions%
\begin{equation}
\mathbf{dexp}_{\hat{\mathbf{X}}}=\sum_{i=0}^{\infty }\frac{1}{\left(
i+1\right) !}\mathbf{ad}_{\hat{\mathbf{X}}}^{i},\ \ \ \ \ \mathbf{dexp}_{%
\hat{\mathbf{X}}}^{-1}=\sum_{i=0}^{\infty }\frac{B_{i}}{i!}\mathrm{ad}_{\hat{%
\mathbf{X}}}^{i}  \label{dexpSeries}
\end{equation}%
where $B_{i}$ are the Bernoulli numbers.

It should be remarked that the body-fixed representation of rigid body
twist, defined by the left Poisson-Darboux equation $\dot{\mathbf{C}}=%
\mathbf{C}\hat{\mathbf{V}}^{\mathrm{b}}$ satisfies the $\mathbf{V}^{\mathrm{b%
}}=\mathbf{dexp}_{-\hat{\mathbf{X}}}\dot{\mathbf{X}}$. Thus, most of the
results presented in this paper can carried over to this representation by
changing the sign of $\mathbf{X}$.

\section{Trajectory Interpolation with Higher-Order Initial Conditions ---
Solution to the $k$-IV-TIP\label{secInterpolInit}}

In the following, $k$th-order accurate formulae are derived that interpolate
the solution of (\ref{Poisson}) for initial value $\mathbf{X}\left( 0\right)
=\mathbf{0}$ and desired terminal value $\mathbf{X}_{T}:=\mathbf{X}\left(
T\right) $ for given initial twist $\mathbf{V}_{0}$ and its derivatives $%
\mathbf{V}^{\left( i\right) },i=1,\ldots ,k-1$.

\subsection{$k$th-Order Solution of the Poisson-Darboux Equation}

Using the Hausdorff-Magnus expansion of the right-trivialized differential,
it was shown in \cite{ZAMM2010} that the solution of the right Poisson
equation can be expressed as a series expansion $\mathbf{X}\left( t\right)
=\sum_{i\geq 0}\frac{t^{i}}{i!}\mathbf{X}_{i}$, where the coefficients are
determined recursively as%
\begin{equation}
\mathbf{X}_{k}=\mathbf{V}_{0}^{\left( k-1\right) }-\left( k-1\right)
!\sum_{j=1}^{k-1}\frac{1}{\left( j-1\right) !}\sum_{l=1}^{k-j}\frac{1}{%
\left( l+1\right) !}\sum_{\pi \in \Pi _{k-j}^{l}}\mathbf{ad}_{\mathbf{Y}%
}^{\pi }\mathbf{X}_{j}  \label{recur2}
\end{equation}%
abbreviating $\mathbf{Y}_{i}=\frac{1}{i!}\mathbf{X}_{i}$, for the sake of
compactness, and $\mathbf{V}_{0}^{\left( k\right) }:=\left. \frac{d^{k}%
\mathbf{V}}{dt^{k}}\right\vert _{t=0}$ is the initial value of the $k$th
derivative of the twist. In (\ref{recur2}) $\Pi _{k}^{l}$ is the set of
ordered partitions of $k$ of length $l$, i.e. $\pi \in \Pi _{k}^{l}$ is an
ordered set of non-negative integers: $\pi =\{\pi _{1},\ldots ,\pi _{l}\}$,
such that $\pi _{1}+\ldots +\pi _{l}=k$. A partition $\pi $ is used as
multidegree so that $\mathrm{ad}_{X}^{\pi }=\mathrm{ad}_{X_{\pi _{1}}}\cdot
\ldots \cdot \mathrm{ad}_{X_{\pi _{l}}}$. A $k$th-order approximation is
therewith obtained as $\mathbf{X}^{\left[ k\right] }\left( t\right)
=\sum_{0\leq i\leq k}\frac{t^{i}}{i!}\mathbf{X}_{i}$. In particular, the
third- and fourth-order approximations are%
\begin{eqnarray}
\mathbf{X}^{[3]}\left( t\right)  &=&t\mathbf{V}_{0}+\frac{1}{2}t^{2}\dot{%
\mathbf{V}}_{0}+\frac{1}{6}t^{3}\ddot{\mathbf{V}}_{0}-\frac{1}{12}t^{3}[%
\mathbf{V}_{0},\dot{\mathbf{V}}_{0}]  \label{approxX3} \\
\mathbf{X}^{[4]}\left( t\right)  &=&t\mathbf{V}_{0}+\frac{1}{2}t^{2}\dot{%
\mathbf{V}}_{0}+\frac{1}{6}t^{3}\ddot{\mathbf{V}}_{0}-\frac{1}{12}t^{3}[%
\mathbf{V}_{0},\dot{\mathbf{V}}_{0}]+\frac{1}{24}t^{4}\dddot{\mathbf{V}}_{0}-%
\frac{1}{24}t^{4}[\mathbf{V}_{0},\ddot{\mathbf{V}}_{0}]  \label{approxX4}
\end{eqnarray}%
where $[\mathbf{X},\mathbf{Y}]=\mathbf{ad}_{\mathbf{X}}\mathbf{Y}$ is the
Lie bracket (also referred to as screw product) of $\mathbf{X},\mathbf{Y}\in
SE\left( 3\right) $, and $\mathbf{ad}_{\mathbf{X}}^{k}\mathbf{Y}=[\mathbf{X}%
,[\mathbf{X},\ldots \lbrack \mathbf{X},\mathbf{Y}]]\ldots ]$ is the $k$-fold
nested Lie bracket.

\subsection{Third-Order Interpolation --- 3-IV-TIP}

The third-order approximation of the screw coordinate at the terminal
configuration, $\mathbf{X}_{T}^{\left[ 3\right] }\approx \log (\mathbf{C}%
_{0}^{-1}\mathbf{C}_{T})$, is obtained with (\ref{approxX3}) as%
\begin{equation}
\mathbf{X}_{T}^{\left[ 3\right] }=T\mathbf{V}_{0}+\frac{1}{2}T^{2}\dot{%
\mathbf{V}}_{0}+\frac{1}{6}T^{3}\ddot{\mathbf{V}}_{0}+\frac{1}{12}T^{3}\left[
\dot{\mathbf{V}}_{0},\mathbf{V}_{0}\right] .  \label{X3T}
\end{equation}%
This can be solved for the highest derivative of $\mathbf{V}_{0}$ as%
\begin{equation}
\ddot{\mathbf{V}}_{0}=\frac{6}{T^{3}}\mathbf{X}_{T}^{\left[ 3\right] }-\frac{%
6}{T^{2}}\mathbf{V}_{0}-\frac{3}{T}\dot{\mathbf{V}}_{0}+\frac{1}{2}\left[ 
\mathbf{V}_{0},\dot{\mathbf{V}}_{0}\right] .  \label{V02d}
\end{equation}%
Enforcing that $\mathbf{X}_{T}^{\left[ 3\right] }$ is the desired value $%
\mathbf{X}_{T}$, and inserting (\ref{V02d}) into the expression (\ref%
{approxX3}) yields%
\begin{equation}
\mathbf{X}^{\left[ 3\right] }\left( \tau \right) =\tau ^{3}\mathbf{X}%
_{T}+\left( \tau -\tau ^{3}\right) \bar{\mathbf{V}}_{0}+\frac{1}{2}\left(
\tau ^{2}-\tau ^{3}\right) \dot{\bar{\mathbf{V}}}_{0}.  \label{X3}
\end{equation}%
The time derivatives of (\ref{X3}) at $t=0$ are $\dot{\mathbf{X}}^{\left[ 3%
\right] }\left( 0\right) =\mathbf{V}_{0}$ and $\ddot{\mathbf{X}}^{\left[ 3%
\right] }\left( 0\right) =\dot{\mathbf{V}}_{0}$. Since at $t=0$, $\dot{%
\mathbf{X}}\left( 0\right) =\mathbf{V}_{0}$ and $\ddot{\mathbf{X}}\left(
0\right) =\dot{\mathbf{V}}_{0}$, the approximation (\ref{X3}) satisfies the
initial conditions. Thus, $\mathbf{C}\left( \tau \right) =\exp (\mathbf{X}^{%
\left[ 3\right] }\left( \tau \right) )\mathbf{C}_{0}$ is a cubic
interpolation formula interpolating the trajectory between initial value $%
\mathbf{X}_{0}=\mathbf{0}$ and terminal value $\mathbf{X}_{T}$, with initial
twist $\mathbf{V}_{0}$ and time derivative $\dot{\mathbf{V}}_{0}$.

\subsection{Fourth-Order Interpolation --- 4-IV-TIP}

Evaluating relation (\ref{approxX4}) at the terminal time yields%
\begin{equation}
\mathbf{X}^{\left( 4\right) }\left( T\right) =T\mathbf{V}_{0}+\frac{1}{2}%
T^{2}\dot{\mathbf{V}}_{0}+\frac{1}{6}T^{3}\ddot{\mathbf{V}}_{0}+\frac{1}{12}%
T^{3}\left[ \dot{\mathbf{V}}_{0},\mathbf{V}_{0}\right] +\frac{1}{24}T^{4}%
\dddot{\mathbf{V}}_{0}+\frac{1}{24}T^{4}\left[ \ddot{\mathbf{V}}_{0},\mathbf{%
V}_{0}\right]  \label{X4T}
\end{equation}%
which delivers a 4th-order approximation $\mathbf{C}\left( \tau \right)
\approx \exp (\mathbf{X}^{\left[ 4\right] }\left( \tau \right) )\mathbf{C}%
_{0}$. This relation is solved as%
\begin{equation}
\dddot{\mathbf{V}}_{0}=\frac{24}{T^{4}}\mathbf{X}_{T}^{\left( 4\right) }-%
\frac{24}{T^{3}}\mathbf{V}_{0}-\frac{12}{T^{2}}\dot{\mathbf{V}}_{0}-\frac{4}{%
T}\ddot{\mathbf{V}}_{0}+\frac{2}{T}\left[ \mathbf{V}_{0},\dot{\mathbf{V}}_{0}%
\right] +\left[ \mathbf{V}_{0},\ddot{\mathbf{V}}_{0}\right] .  \label{V03d}
\end{equation}%
Expressed in terms of the prescribed terminal value $\mathbf{X}_{T}$, and
inserting this in (\ref{approxX4}) yields%
\begin{equation}
\mathbf{X}^{\left( 4\right) }\left( \tau \right) =\tau ^{4}\mathbf{X}%
_{T}+\left( \tau -\tau ^{4}\right) \bar{\mathbf{V}}_{0}+\frac{1}{2}\left(
\tau ^{2}-\tau ^{4}\right) \dot{\bar{\mathbf{V}}}_{0}+\frac{1}{6}\left( \tau
^{3}-\tau ^{4}\right) (\ddot{\bar{\mathbf{V}}}_{0}+\frac{1}{2}[\dot{\bar{%
\mathbf{V}}}_{0},\bar{\mathbf{V}}_{0}]).  \label{X4}
\end{equation}%
The time derivatives of (\ref{X4}) at $t=0$ are $\dot{\mathbf{X}}^{\left[ 3%
\right] }\left( 0\right) =\mathbf{V}_{0},\ddot{\mathbf{X}}\left( 0\right) =%
\dot{\mathbf{V}}_{0}$, and $\dddot{\mathbf{X}}^{\left[ 3\right] }\left(
0\right) =\ddot{\mathbf{V}}_{0}$. Clearly, for initial twist and its
derivatives prescribed to be zero, both (\ref{X3}) and (\ref{X4}), describe
the geodesic motion $\mathbf{C}\left( t\right) =\exp \left( \mathbf{X}%
_{T}t\right) \mathbf{C}_{0}$, but with a trajectory depending on time
according to $\tau ^{3}$ and $\tau ^{4}$, respectively.

\section{Higher-Order Trajectory Interpolation with Higher-Order Initial and
Terminal Conditions}

The above formulae for interpolating spatial motions between initial and
terminal configuration, described by $\mathbf{X}\left( T\right) $, admit
specifying the initial twist and derivatives. They do not allow for
prescribing the twist, and its derivatives if desired, at the terminal
configuration.

\subsection{Cubic Trajectory Interpolation}

A cubic interpolation that allows prescribing the terminal twist can be
derived from (\ref{X3}). To this end, $\dot{\mathbf{V}}_{0}$ is related to
the terminal twist. The twist at terminal time $T$ is expressed using (\ref%
{RecExp}) as $\mathbf{V}\left( T\right) =\mathbf{dexp}_{\hat{\mathbf{X}}_{T}}%
\dot{\mathbf{X}}_{T}$. The screw coordinate vector at terminal time $T$ are
obtained from the cubic interpolation formula (\ref{X3}). Its time
derivative is%
\begin{equation}
\dot{\mathbf{X}}_{T}=\frac{1}{T}\left( 3\mathbf{X}_{T}-2\bar{\mathbf{V}}_{0}-%
\frac{1}{2}\dot{\bar{\mathbf{V}}}_{0}\right)  \label{X3dT}
\end{equation}%
and thus%
\begin{equation}
\bar{\mathbf{V}}_{T}=\mathbf{dexp}_{\hat{\mathbf{X}}_{T}}\left( 3\mathbf{X}%
_{T}-2\bar{\mathbf{V}}_{0}-\frac{1}{2}\dot{\bar{\mathbf{V}}}_{0}\right) .
\label{V3T}
\end{equation}%
Solving (\ref{V3T}) for $\dot{\bar{\mathbf{V}}}_{0}$ and inserting this into
(\ref{X3}) yields%
\begin{equation}
\mathbf{X}^{\left[ 3\right] }\left( \tau \right) =\left( 3\tau ^{2}-2\tau
^{3}\right) \mathbf{X}_{T}+\tau \left( 1-\tau \right) ^{2}\bar{\mathbf{V}}%
_{0}+\tau ^{2}\left( \tau -1\right) \mathbf{dexp}_{\hat{\mathbf{X}}_{T}}^{-1}%
\bar{\mathbf{V}}_{T}.  \label{CubicInterp}
\end{equation}

Formula (\ref{CubicInterp}) is a novel cubic interpolation between spatial
configurations with given initial and terminal twist. It should be remarked
that, if initial and terminal twist are zero, $\mathbf{X}^{\left[ 3\right]
}\left( \tau \right) $ defined in (\ref{CubicInterp}) yields the minimal
acceleration curve as shown by Zefran \cite{ZefranKumarCroke-TRO1998}, which
is also a geodesic on $SE\left( 3\right) $ w.r.t. to the left invariant
metric on $SE\left( 3\right) $. A similar formula was presented in \cite%
{ParkRavani1997,KangPark1998}, where the point of departure was a general
cubic approximation of $\mathbf{X}\left( \tau \right) $, rather than one
arising from an approximate solution as in section \ref{secInterpolInit}.
The trajectory obtained from the interpolation proposed in \cite%
{ParkRavani1997,KangPark1998} does not yield the minimal acceleration curve.

\subsection{Higher-Order Trajectory Interpolation}

Further higher-order interpolation schemes can be derived following the
approach above. To this end, the derivatives of the twist at initial time
are replaced by those the terminal time. The order of the interpolation from
section \ref{secInterpolInit} must be chosen according to the number of
derivatives prescribed at terminal time. This replacement necessitates the
time derivatives of (\ref{Xdot}), and thus of the dexp map. Closed form
expressions are available \cite{RSPA2021} for the first derivative of dexp
on $SE\left( 3\right) $ and for the second derivative on $SO\left( 3\right) $%
. Higher-order derivatives in closed form have not yet been published, and
will be reported in a forthcoming paper.%
\newpage%

\section{Examples}

The cubic interpolation formula (\ref{CubicInterp}) is applied to
interpolate between two rigid body configurations along a prescribed
trajectory described by $\mathbf{C}\left( t\right) =\exp \mathbf{X}\left(
t\right) $ with known $\mathbf{X}\left( t\right) $. Two motions, described
by $\mathbf{C}_{1}$ and $\mathbf{C}_{2}$, are compared using the norm $||%
\mathbf{C}_{1}^{-1}\mathbf{C}_{2}||:=||\log (\mathbf{C}_{1}^{-1}\mathbf{C}%
_{2})||=\alpha \left\Vert \mathbf{x}\right\Vert +\beta \left\Vert \mathbf{y}%
\right\Vert $, induced by the left-invariant metric on $SE\left( 3\right) $,
of the relative configuration%
\begin{equation}
\mathbf{C}_{1}^{-1}\mathbf{C}_{2}=\left( 
\begin{array}{cc}
\mathbf{R} & \mathbf{r} \\ 
\mathbf{0} & 1%
\end{array}%
\right)
\end{equation}%
where $\mathbf{x}=\log (\mathbf{R}_{1}^{-1}\mathbf{R}_{2})$ is the logarithm
on $SO\left( 3\right) $, and $\mathbf{y}=\mathbf{dexp}_{\mathbf{x}}^{-1}%
\mathbf{r}$, with the $\mathbf{dexp}$ map on $SO\left( 3\right) $. To avoid
ambiguity, due to the scale dependence of the metric, the terms $\left\Vert 
\mathbf{x}\right\Vert $ and $\left\Vert \mathbf{y}\right\Vert $ will be
computed separately.

\subsection{Motion Cubic in Time}

\begin{figure}[b]
\begin{centering}
a)~~\includegraphics[width=8.7cm]{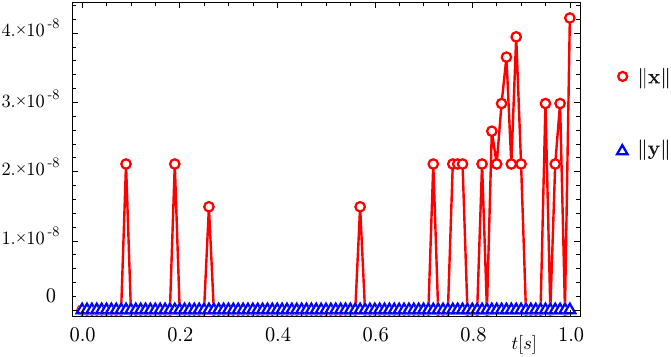}\\
b)~~\includegraphics[width=8cm]{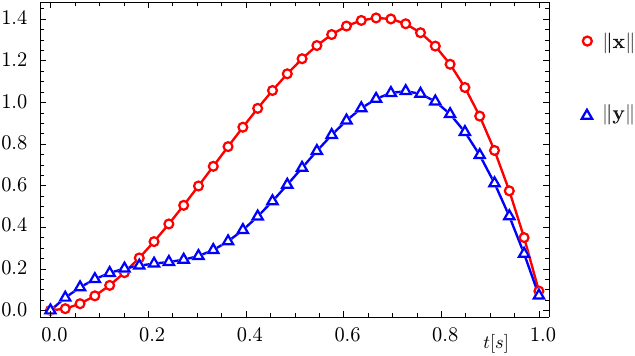}\\
\end{centering}
\caption{Difference $||\mathbf{C}^{-1}\mathbf{C}^{\left[ 3\right] }||$ of
the cubic approximation $\mathbf{C}^{\left[ 3\right] }\left( t\right) $ and
the actual motion $\mathbf{C}\left( t\right) $. a) When complete initial and
terminal state is provided. b) When initial and terminal twists are zero. }
\label{figCubic}
\end{figure}
\begin{figure}[th]
\begin{centering}
\includegraphics[width=8cm]{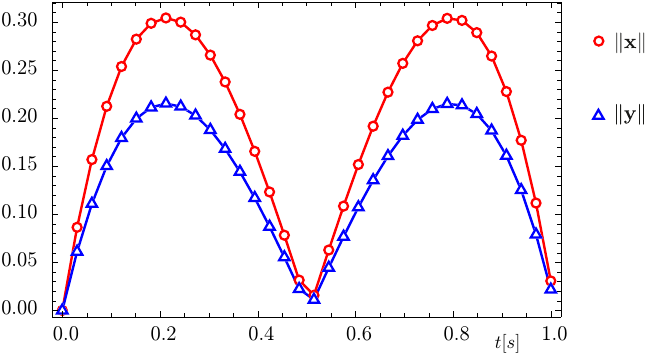}\\
\end{centering}
\caption{Difference $||\mathbf{C}_{\mathrm{geo}}^{-1}\mathbf{C}^{\left[ 3%
\right] }||$ of the geodesic and the mimimun accleration trajectory.}
\label{figDiff_MinAcc-Geod_Cubic}
\end{figure}

The rigid motion is prescribed with $\mathbf{X}\left( t\right) =\left(
0,3t^{3},t^{3},2t,0,t\right) $. An interpolation is computed from initial
pose at $t=0$ to terminal pose $T=1$. The latter is used for simplicity, so
that $t=\tau $. The initial and terminal twist are $\mathbf{V}_{0}=\left(
0,0,0,2,0,1\right) $ and $\mathbf{V}_{T}=\left(
0,9,3,4.82629,-1.40384,5.21152\right) $, respectively.

First, the interpolation $\mathbf{C}^{\left[ 3\right] }\left( t\right) =\exp 
\mathbf{X}^{\left[ 3\right] }\left( t\right) $ is computed with (\ref%
{CubicInterp}) where the terminal pose $\mathbf{X}_{T}=\left(
0,3,1,2,0,1\right) $ as well as the initial terminal twists are supplied.
The cubic interpolation formula exactly reproduces the motion. Fig. \ref%
{figCubic}a) shows the error $||\mathbf{C}^{-1}\mathbf{C}^{\left[ 3\right]
}||$ along the trajectory. Notice, that the interpolation achieves a perfect
match for any trajectory that is polynomial in time up to order 3. Next, the
trajectory $\mathbf{C}^{\left[ 3\right] }\left( t\right) $ is computed with
zero initial and terminal twists, $\mathbf{V}_{0}=\mathbf{V}_{T}=\mathbf{0}$%
. This yields the minimum acceleration trajectory. The result is shown in
Fig. \ref{figCubic}b). Setting the twists to zero clearly corresponds to a
different trajectory than the one prescribed from which the initial
configurations are deduced. However, if initial/terminal twists are not
prescribed, the minimum acceleration curve is a sensible choice for motion
planning that was treated in the literature \cite%
{ZefranKumar1998,ParkRavani1997,KangPark1998,Selig-IMA2007}. The geometric
path of the minimum acceleration trajectory is also a geodesic. It is
instructive to compare this with the geodesic obtained from the simple
exponential linear interpolation $\mathbf{C}_{\mathrm{geo}}\left( t\right)
=\exp (\mathbf{X}_{T}t)$. Fig. \ref{figDiff_MinAcc-Geod_Cubic} shows the
difference $||\mathbf{C}_{\mathrm{geo}}^{-1}\mathbf{C}^{\left[ 3\right] }||$%
, where $\mathbf{C}^{\left[ 3\right] }\left( t\right) $ is the minimum
acceleration curve. It can be seen that there is a significant difference of
the motions although both follow the same geometric path.

\subsection{General Motion}

\begin{figure}[t]
\begin{centering}
a)~~\includegraphics[width=8.5cm]{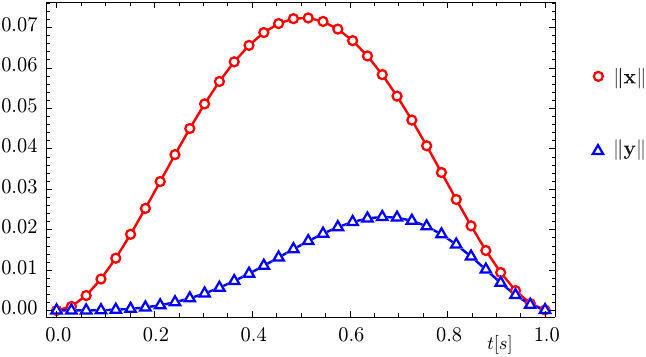}\\
b)~~\includegraphics[width=8.5cm]{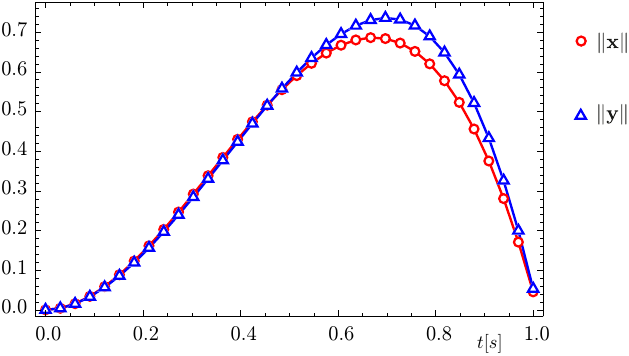}\\
\end{centering}
\caption{Difference $||\mathbf{C}^{-1}\mathbf{C}^{\left[ 3\right] }||$ of
the cubic approximation $\mathbf{C}^{\left[ 3\right] }\left( t\right) $ and
the actual motion $\mathbf{C}\left( t\right) $. a) When complete initial and
terminal state is provided. b) When initial and terminal twists are zero. }
\label{figQuartic}
\end{figure}
\begin{figure}[th]
\begin{centering}
\includegraphics[width=8.5cm]{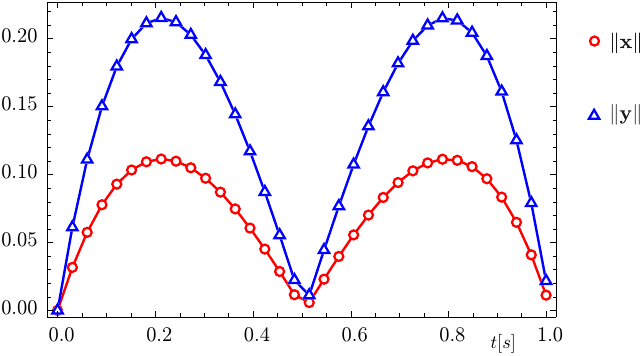}\\
\end{centering}
\caption{Difference $||\mathbf{C}_{\mathrm{geo}}^{-1}\mathbf{C}^{\left[ 3%
\right] }||$ of the geodesic and the mimimun accleration trajectory.}
\label{figDiff_MinAcc-Geod_Quartic}
\end{figure}
As a more general motion to be interpolated, a trajectory described by a
fourth-order polynomial in time is considered, which is described by the
screw coordinate vector $\mathbf{X}\left( t\right) =\left(
-t^{4},0.3t^{4},0.5t^{4},2t^{2},0,t^{2}\right) $. It is clear from Fig. \ref%
{figQuartic}a) that the interpolation does not perfectly match the original
trajectory, which cannot be expected from an interpolation scheme. If
initial and terminal twist are set to zero, this difference become larger as
shown in \ref{figQuartic}b). Yet the so obtained geodesic is excellent
interpolation. The difference of this minimum acceleration curve and the
geodesic $\mathbf{C}_{\mathrm{geo}}\left( t\right) $ is shown in Fig \ref%
{figQuartic}c).%
\newpage%

\ 
\newpage%

\section{Conclusion and Outlook}

The interpolation of rigid body motion has been revisited. Interpolation
formulae are derived for when the initial velocity and its derivatives are
prescribed, and for the case when this is prescribed at the initial and
terminal pose. The presented expressions can be straightforwardly
generalized to account for prescribed time derivatives of arbitrary degree.
In particular, a 4th-order expression analogous to (\ref{CubicInterp}) can
be derived. This gives rise to a 4th-order (i.e. quartic instead of cubic)
spline interpolation scheme. Further higher-order splines can be derived
that allow 1) prescribing the time derivatives of the twist at the knot
point, or 2) to compute splines with higher-order compatibility at the knot
points. The interpolation is not restricted to interpolating time dependent
motions. It can be used to interpolate the deformation field of
geometrically (Cosserat) exact beams. Then, $\tau $ is the arc length, the
beam section is represented by $\mathbf{C}\left( \tau \right) $, and the
beam kinematics is described by the equation $\mathbf{C}^{\prime }=\hat{%
\bm{\chi}}{^{\mathrm{s}}}\mathbf{C}$, where the strain field is defined by
the deformation measure $\hat{\bm{\chi}}:\left[ 0,L\right] \rightarrow
se\left( 3\right) $, also called base-pole generalized curvature \cite%
{BorriBottasso1994a,BorriBottasso1994b}. Recent progress in Cosserat beam
modeling used the left-invariant curvature $\hat{\bm{\chi}}{^{\mathrm{b}}}$
defined by $\mathbf{C}^{\prime }=\mathbf{C}\hat{\bm{\chi}}{^{\mathrm{b}}}$ 
\cite{SonnevilleCardonaBruls2014}.

The formulae presented in section \ref{secInterpolInit} can be applied in
higher-order model predictive control (MPC) for spatial guidance control.

The interpolation scheme yields the trajectory $\mathbf{X}\left( t\right) $
of the canonical coordinates, i.e. a curve in $se\left( 3\right) $. Instead
of using $SE\left( 3\right) $, the corresponding motion can be represented
in any Lie group whose Lie algebra is isomorphic to $se\left( 3\right) $. As
such dual quaternions, forming the Lie group $\bar{Sp}\left( 1\right) $ can
be used \cite{CND2016}. Alternative interpolation formulae can be derived in
terms of non-canonical coordinates, e.g. in terms if generalized Rodrigues
parameters using the Cayley map on $SE\left( 3\right) $.

\section*{Acknowledgement}

This work has been supported by the LCM K2 Center for Symbiotic Mechatronics
within the framework of the Austrian COMET-K2 program.

\bibliographystyle{spmpsci}
\bibliography{ExpCaySE3}

\begin{thebibliography}{10}
\providecommand{\url}[1]{{#1}}
\providecommand{\urlprefix}{URL }
\expandafter\ifx\csname urlstyle\endcsname\relax
  \providecommand{\doi}[1]{DOI~\discretionary{}{}{}#1}\else
  \providecommand{\doi}{DOI~\discretionary{}{}{}\begingroup
  \urlstyle{rm}\Url}\fi

\bibitem{BorriBottasso1994a}
Borri, M., Bottasso, C.: {An intrinsic beam model based on a helicoidal
  approximation—Part I: Formulation}.
\newblock International Journal for Numerical Methods in Engineering
  \textbf{37}(13), 2267--2289 (1994)

\bibitem{BorriBottasso1994b}
Borri, M., Bottasso, C.: {An intrinsic beam model based on a helicoidal
  approximation—Part II: Linearization and finite element implementation}.
\newblock International journal for numerical methods in engineering
  \textbf{37}(13), 2291--2309 (1994)

\bibitem{Hausdorff1906}
Hausdorff, F.: {Die symbolische {E}xponentialformel in der {G}ruppentheorie}.
\newblock Berichte der K{\"o}niglich-S{\"a}chsischen Geselschaft der
  Wissenschaften zu Leipzig, Mathematisch-Physische Klasse \textbf{58}, 19--48
  (1906)

\bibitem{Iserles1984}
Iserles, A.: Solving linear ordinary differential equations by exponentials of
  iterated commutators.
\newblock Numerische Mathematik \textbf{45}(2), 183--199 (1984)

\bibitem{KangPark1998}
Kang, I., Park, F.: Cubic spline algorithms for orientation interpolation.
\newblock International journal for numerical methods in engineering
  \textbf{46}(1), 45--64 (1999)

\bibitem{Magnus1954}
Magnus, W.: On the exponential solution of differential equations for a linear
  operator.
\newblock Communications on pure and applied mathematics \textbf{7}(4),
  649--673 (1954)

\bibitem{ZAMM2010}
M{\"u}ller, A.: Approximation of finite rigid body motions from velocity
  fields.
\newblock ZAMM-Journal of Applied Mathematics and Mechanics/Zeitschrift f{\"u}r
  Angewandte Mathematik und Mechanik: Applied Mathematics and Mechanics
  \textbf{90}(6), 514--521 (2010)

\bibitem{CND2016}
M\"{u}ller, A.: {Coordinate mappings for rigid body motions}.
\newblock ASME Journal of Computational and Nonlinear Dynamic \textbf{12(2)}
  (2016)

\bibitem{RSPA2021}
M{\"u}ller, A.: {Review of the exponential and Cayley map on SE (3) as relevant
  for Lie group integration of the generalized Poisson equation and flexible
  multibody systems}.
\newblock Proceedings of the Royal Society A \textbf{477}(2253) (2021)

\bibitem{Noakes1989}
Noakes, L., Heinzinger, G., Paden, B.: Cubic splines on curved spaces.
\newblock IMA Journal of Mathematical Control and Information \textbf{6}(4),
  465--473 (1989)

\bibitem{ParkRavani-JMD1995}
Park, F., Ravani, B.: {B{\'{e}}zier curves on Riemannian manifolds and Lie
  groups with kinematics applications}.
\newblock J. of Mechanical Design \textbf{117}(1), 36--40 (1995)

\bibitem{ParkRavani1997}
Park, F.C., Ravani, B.: Smooth invariant interpolation of rotations.
\newblock ACM Transactions on Graphics (TOG) \textbf{16}(3), 277--295 (1997)

\bibitem{Selig-IMA2007}
Selig, J.: {Curves of stationary acceleration in SE(3)}.
\newblock IMA Journal of Mathematical Control and Information \textbf{24}(1),
  95--113 (2007)

\bibitem{Selig-IFToMM2007}
Selig, J.M.: {Cayley maps for SE(3)}.
\newblock In: 12th International Federation for the Promotion of Mechanism and
  Machine Science World Congress, p.~6 (2007)

\bibitem{Selig-IROS2006}
Selig, J.M., Wu, Y.: Interpolated rigid-body motions and robotics.
\newblock In: 2006 IEEE/RSJ International Conference on Intelligent Robots and
  Systems, pp. 1086--1091. IEEE (2006)

\bibitem{Shoemake1985}
Shoemake, K.: Animating rotation with quaternion curves.
\newblock In: Proceedings of the 12th annual conference on Computer graphics
  and interactive techniques, pp. 245--254 (1985)

\bibitem{SonnevilleCardonaBruls2014}
Sonneville, V., Cardona, A., Br{\"u}ls, O.: {Geometrically exact beam finite
  element formulated on the special Euclidean group SE(3)}.
\newblock Computer Methods in Applied Mechanics and Engineering \textbf{268},
  451--474 (2014)

\bibitem{ZefranKumar1998}
{\v{Z}}efran, M., Kumar, V.: Interpolation schemes for rigid body motions.
\newblock Computer-Aided Design \textbf{30}(3), 179--189 (1998)

\bibitem{ZefranKumarCroke-TRO1998}
{\v{Z}}efran, M., Kumar, V., Croke, C.B.: On the generation of smooth
  three-dimensional rigid body motions.
\newblock IEEE Transactions on Robotics and Automation \textbf{14}(4), 576--589
  (1998)

\end{thebibliography}

\end{document}